\documentclass[letterpaper]{article} 
\usepackage{aaai23}  
\usepackage{times}  
\usepackage{helvet}  
\usepackage{courier}  
\usepackage[hyphens]{url}  
\usepackage{graphicx} 
\usepackage{natbib}  
\usepackage{caption} 
\usepackage{amsmath,amssymb,amsfonts}
\usepackage{array}
\usepackage{textcomp}

\usepackage{stfloats}
\usepackage{url}
\usepackage{breakurl}
\usepackage{textcomp}
\usepackage{booktabs}

\usepackage{longtable}
\usepackage{multirow}
\usepackage{multicol}
\usepackage{footnote}
\usepackage{mathrsfs}
\usepackage{threeparttable}

\usepackage{algorithm}
\usepackage{algorithmic}

\usepackage{newfloat}
\usepackage{listings}
\DeclareCaptionStyle{ruled}{labelfont=normalfont,labelsep=colon,strut=off} 
\floatstyle{ruled}
\newfloat{listing}{tb}{lst}{}
\floatname{listing}{Listing}
\title{Spatio-temporal Neural Structural Causal Models for Bike Flow Prediction}
\author {
    Pan Deng\textsuperscript{\rm 1},
    Yu Zhao\textsuperscript{\rm 1}\thanks{Corresponding author.},
    Junting Liu\textsuperscript{\rm 1},
    Xiaofeng Jia\textsuperscript{\rm 2},
    Mulan Wang\textsuperscript{\rm 1}
}
\affiliations {
    \textsuperscript{\rm 1} Beihang University, Beijing, 100191, China.\\
    \textsuperscript{\rm 2} Beijing Big Data Centre, Beijing, 100024, China.\\
     \{pandeng, iyzhao, liujunting, wangmulan\}@buaa.edu.cn,   jiaxf@jxj.beijing.gov.cn
}

\usepackage{bibentry}

\begin{document}

\maketitle

\begin{abstract}
As a representative of public transportation, the fundamental issue of managing bike-sharing systems is bike flow prediction. Recent methods overemphasize the spatio-temporal correlations in the data, ignoring the effects of contextual conditions on the transportation system and the inter-regional time-varying causality.
In addition, due to the disturbance of incomplete observations in the data, random contextual conditions lead to spurious correlations between data and features, making the prediction of the model ineffective in special scenarios. To overcome this issue, we propose a Spatio-temporal Neural Structure Causal Model(STNSCM) from the perspective of causality. First, we build a causal graph to describe the traffic prediction, and further analyze the causal relationship between the input data, contextual conditions, spatio-temporal states, and prediction results. Second, we propose to apply the frontdoor criterion to eliminate confounding biases in the feature extraction process. Finally, we propose a counterfactual representation reasoning module to extrapolate the spatio-temporal state under the factual scenario to future counterfactual scenarios to improve the prediction performance. Experiments on real-world datasets demonstrate the superior performance of our model, especially its resistance to fluctuations caused by the external environment. The source code and data will be released.

\end{abstract}

\section{Introduction}
\label{sec:introduction}

Bike-Sharing systems have been widely deployed in urban public transportation due to their convenience and environmental friendliness in recent years.
As a representative of Intelligent Transportation System (ITS), one of the key concerns is the effective allocation of bike-sharing resources to enhance the quality of system service. However, due to the high frequency and randomness of bike usage throughout the city, the stations often get imbalanced over time. The bike-sharing system always has some congested and hungry stations, leading to a significant number of unsatisfied customers. Therefore, it is necessary to accurately predict the bike flow in each commuting area. According to the prediction results of bike flow, it can be repositioned in advance to prevent excessive demand at the station, which is an urgent need for bike-sharing system operators.

\begin{figure}
  \centering
  \includegraphics[width=0.5\textwidth]{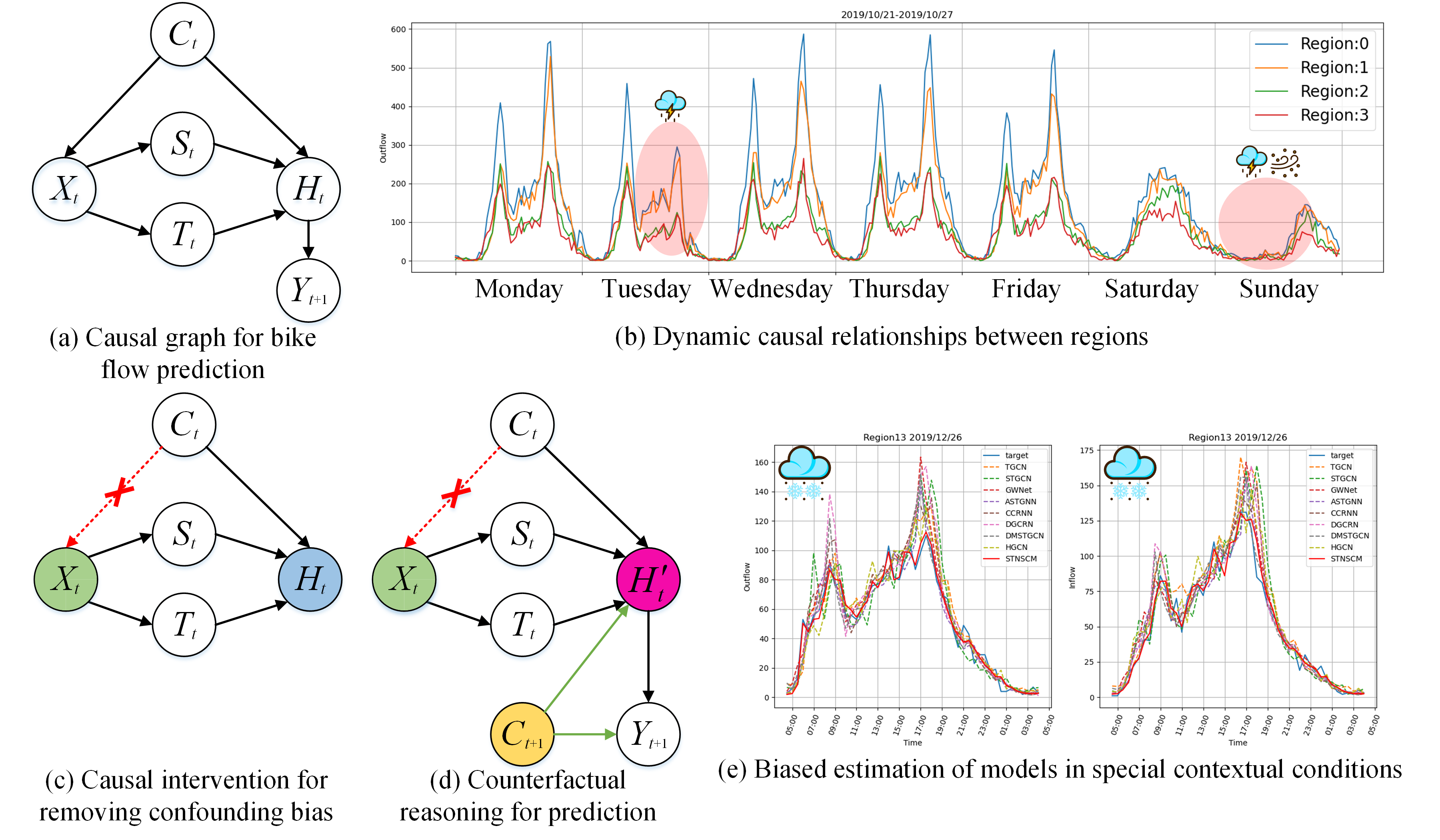}
  \caption{The change process of bike flow from the perspective of causality.}
  \label{fig:1}
\end{figure}

Bike usage patterns and spatio-temporal dependencies between regions are affected by external conditions (e.g. time factors and weather information). As shown in Fig.\ref{fig:1}(b), the time factors and weather may greatly restrict bike usage in each region, and weekdays and weekends may cause completely different spatio-temporal dependencies. Traditional traffic flow prediction models based on multi-source data usually simply integrate time and weather into the model through the fully connected layer\cite{li2019learning, liang2021fine, sun2020predicting}. More input information cannot improve the prediction ability of the model. Instead, it will introduce a large number of confounding factors and extract spurious correlations in observations\citep{liu2021learning}, resulting in the decline of model performance.

We build a causal graph to describe the bike flow prediction, and further analyze it from a causal perspective. We formulate the causalities among bike flow $\mathbf{X}_t$, contextual condition $\mathbf{C}_t$, spatio-temporal states $\mathbf{H}_t$, and predicted target $\mathbf{Y}_{t+1}$. As shown in Fig. \ref{fig:1}(a), directed edges denote causal relationships between nodes. The spatio-temporal state $\mathbf{H}_t$ represents the intrinsic trend of bicycle demand under the current contextual conditions. The contextual condition $\mathbf{C}_t$, as a common cause, affects $\mathbf{X}_t$ and the spatio-temporal state $\mathbf{H}_t$. Bike-sharing systems are susceptible to external contextual conditions, so bike flow  prediction must take contextual conditions into account. However, due to the limitation of dataset, the data are collected under normal conditions in most cases, which causes the neural network to only learn the general spatio-temporal patterns under normal conditions, and it is difficult to capture the spatio-temporal state under special conditions. Once a special contextual condition occurs during testing, the predictive performance of the model will drop significantly. As shown in Fig. \ref{fig:1}(e), there is a strong snowstorm on this day, the flow is generally low and the rush hours in the morning and evening are not obvious, which is a special environment that unseens in the training set. The predictions of existing methods are on the high side, because they learn a more average spatio-temporal pattern, which is difficult to cope with fluctuations caused by the external environment. These methods emphasize the average performance over the entire dataset while ignoring the prediction performance in specific scenarios. However, bike flow prediction in specific scenarios are often more helpful for managers to formulate emergency measures in advance. Therefore, it is necessary to eliminate the influence of potential  contextual conditions on feature extraction by means of causal intervention, so as to make the extracted spatio-temporal state more fair and effectively capture the spatio-temporal pattern of the data itself.

In addition, recent works overemphasize spatio-temporal correlations of bike flow. Generally, these graph-based spatio-temporal prediction models leverage GCN to model spatial correlations, and GRU\cite{li2018dcrnn, liu2020physical, NIPS2020AGCRN, ye2021CGCdemand, li2021DGCRN} or CNN\cite{graphwavenet2019, AAAI2021HGCN, KDD2021ODEGCN, KDD2021DMSTGCN} to model temporal correlations separately. Although these methods can achieve satisfactory effects, their ability to model complex nonlinear dynamic spatio-temporal causality is still obviously insufficient. As shown in Fig.\ref{fig:1}(b), bike usage patterns in region 1 are similar to region 2 and 3 during the morning rush hour, but are more similar to region 0 during the evening rush hour. Therefore, the bike flow data implies intense dynamic dependencies in spatial and temporal dimensions and complex nonlinear causality in spatio-temporal dimensions. Most of the adjacency matrices are fixed and generated by heuristic methods based on spatial distance\cite{liu2020physical} or time series similarity\cite{bikemultigraph2018}, which cannot capture the time-varying spatio-temporal causality.

To address these aforementioned challenges, We propose a Spatio-Temporal Neural Structural Causal Model (STNSCM), and the causal graph shown in Fig. \ref{fig:1}(c) and (d). Its core idea is based on structural causal model theory to remove confounders in the feature extraction process and follow the counterfactual reasoning framework to predict future bike flow. First, we apply the frontdoor criterion based on causal intervention, cutting off the link $\mathbf{C}_t \to \mathbf{X}_t$, which gives $\mathbf{X}_t$ a fair opportunity to incorporate each contextual condition $\mathbf{C}$ into spatio-temporal state $\mathbf{H}_t$. Second, we view future scenarios in a "what if" way, that is, if the current environment changes, how will the future state change, and thus how will future flow change? The key to answer this counterfactual question is how to make full use of future external conditions. Specifically, The main contributions are as follows:

\begin{itemize}

\item We provide a novel causality-based interpretation for the bike flow prediction and apply the frontdoor criterion based on causal interventions to remove confounding biases in the feature extraction process. To the best of our knowledge, our work is the first one that successfully applies the structural causal model to traffic prediction problems.

\item We propose a counterfactual representation reasoning module to extrapolate the spatio-temporal state under the factual scenario to the future counterfactual scenario, which enhances the feature's understanding of future states, thereby improving the prediction performance.




\item Extensive experiments on two real-world bike-sharing systems datasets show that our STNSCM comprehensively outperforms state-of-the-art methods for region-level bike flow prediction.

\end{itemize}

\section{Related Work}

\subsection{Spatio-temporal Traffic Data Prediction}

Traffic prediction is a fundamental problem in Intelligent Transportation System including the recent Bike-Sharing System, which has recently attracted significant attention. Since the real-world traffic network is non-Euclidean data, most methods construct GCN models to study spatio-temporal data prediction problems. utilized GCN with predefined graphs to model spatial correlations. To enrich spatial information, \cite{liu2020physical, KDD2021ODEGCN, bikemultigraph2018} establish multiple static topologies to effectively capture complex patterns. DCRNN\cite{li2018dcrnn} replaces the linear transformation layer in GRU with diffusion graph convolution, which boosts the spatio-temporal representation ability of GRU. HGCN\cite{AAAI2021HGCN} constructs the interaction between the micro and macro layers of GCN, which integrates the different scales of features of road segments and regions.

Recently, some researchers have paid attention to the strong dynamic correlations of traffic data in spatial and temporal dimensions. Therefore, it is crucial to model dynamic and nonlinear spatio-temporal correlations for accurate traffic prediction. GWNet\cite{graphwavenet2019} and AGCRN\cite{NIPS2020AGCRN} randomly initialize the node embedding vectors, and then learn an adaptive matrix through gradient descent. Once the training of the model finishes, the adaptive matrix is fixed. So it cannot extract the dynamics from real data effectively. ASTGNN\cite{TKDE2021ASTGNN} and GMAN\cite{AAAI2020GMAN} dynamically extract the correlations of nodes by means of spatial attention. However, affected by the structure of predefined graphs, spatial attention can only change the weight of predefined graphs rather than the structure. CCRNN\cite{ye2021CGCdemand} proposes a coupled graph convolution with self-learned adjacency matrices varying from layer to layer, but the graph structure changes with the convolutional layer instead of time steps. DGCRN\cite{li2021DGCRN} handles the dynamic relations by learning the matrix at each recurrent step, while it significantly depends on the node embedding layer initialized by random parameters. DMSTGCN\cite{KDD2021DMSTGCN} learns the time-specific spatial dependencies of road segments to construct dynamic graphs, but ignores the fluctuations caused by the external environment. In addition, these methods do not take external conditions into consideration during the dynamic graph generation process, resulting in very limited effects of dynamic graphs.

\subsection{Causal learning}

The purpose of causal learning is to empower models the ability to pursue the causal effect \citep{zhang2020causal}. Recent works are proposed to combine SCM \citep{pearl2009causality, scholkopf2022causality} with deep learning models, which is called causal learning. Causal learning is widely used in the field of causal representation, CasualVAE \citep{yang2021causalvae} proposes a model with causal layer to transform exogenous factors into causal endogenous ones that correspond to causally related concepts in data. Shen \citep{shen2020disentangled}et al. use a SCM as the prior for bidirectional generative model which can generate data from any desired interventional distributions of the latent factors. \citep{zhang2020causal, Lin_Chen_Li_Yu_2022, liu2022towards, yue2020interventional} use the backdoor criterion to eliminate the contextual confounding biases. Different from above works, the input data and extracted features do not satisfy the backdoor criterion in the field of traffic prediction. As shown in Fig. \ref{fig:1}(a), we apply the frontdoor formula and use neural networks to model the sub terms, so our model is called neural structural causal model.

\section{Methodology}

\begin{figure*}
  \centering
  \includegraphics[width=1\textwidth]{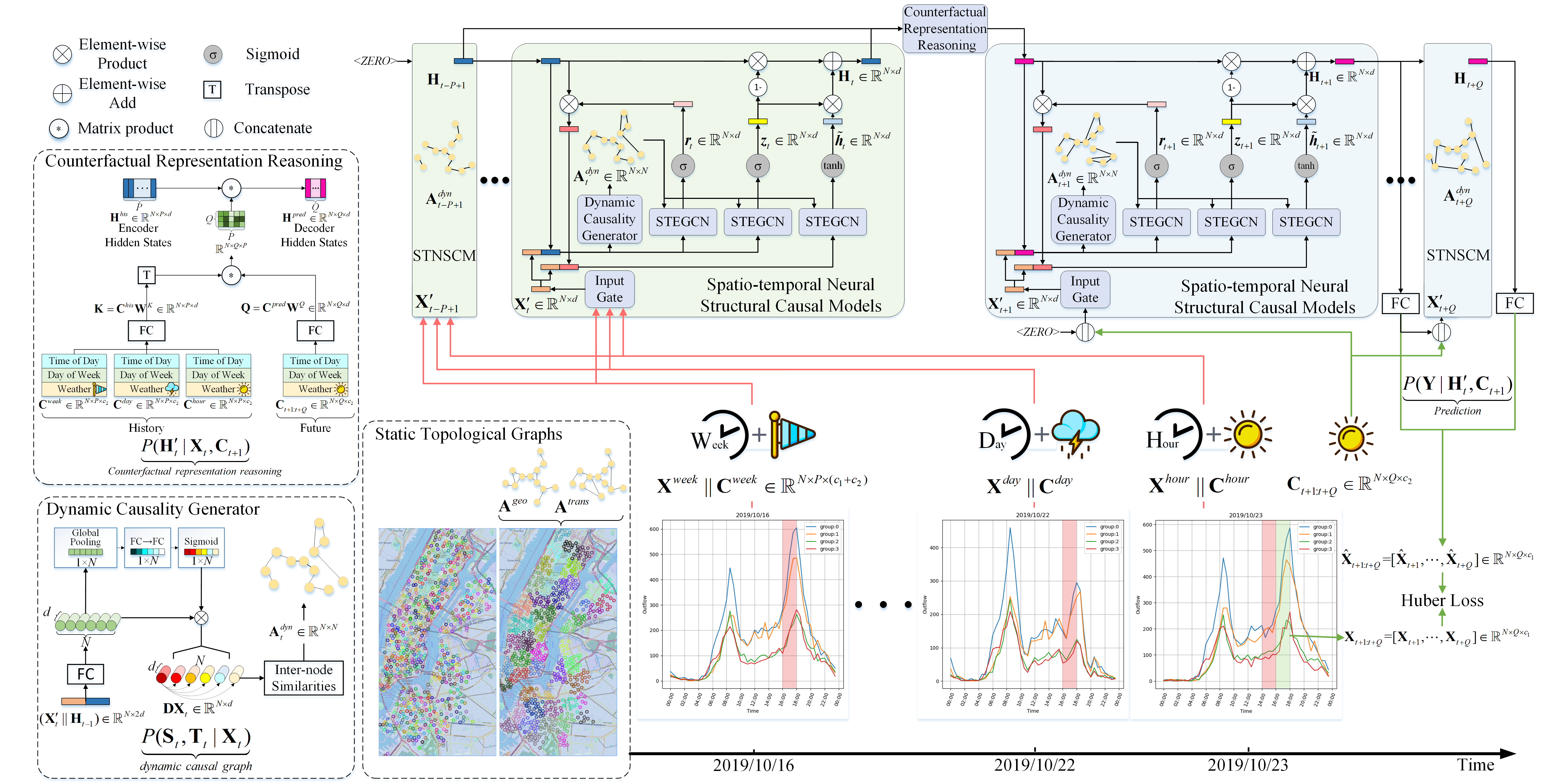}
  \caption{The architecture of STNSCM.}
  \label{fig:2}
\end{figure*}

The overall framework of STNSCM is shown in Fig.\ref{fig:2}.

\subsection{Structural Causal Model}
We formulate the causalities among bike flow $\mathbf{X}_t$, contextual condition $\mathbf{C}_t$, spatial neighborhood features $\mathbf{S}_t$, temporal dynamic features $\mathbf{T}_t$, spatio-temporal states $\mathbf{H}_t$, and predicted target $\mathbf{Y}_{t+1}$, with a structural Causal Model (SCM). As shown in Fig. \ref{fig:1}(a), directed edges denote causal relationships between nodes. We believe that spatial neighborhood features $\mathbf{S}_t$ and temporal dynamic features $\mathbf{T}_t$ can be decoupled from spatio-temporal data $\mathbf{X}_t$, and they are integrated to form spatiotemporal states $\mathbf{H}_t$, which can describe dynamic spatio-temporal patterns in the data.

\subsection{Causal Intervention via Frontdoor Criterion}

The contextual condition $\mathbf{C}_t$ is expressed as the common cause of $\mathbf{X}_t$ and $\mathbf{H}_t$, which may cause $\mathbf{H}_t$ to be more inclined to the general state and ignore the specific environment due to the limitation of the dataset, resulting in unfair bias of $\mathbf{H}_t$. However, the $\mathbf{C}_t$ have infinite possibilities and cannot be completely covered. we cannot use backdoor adjustments according to the path $\mathbf{X}_t \gets  \mathbf{C}_t \to \mathbf{H}_t$.

Fortunately, we apply the frontdoor criterion based on the path $\mathbf{C}_t \to \mathbf{X}_t \to  \mathbf{S}_t, \mathbf{T}_t \to \mathbf{H}_t \gets \mathbf{C}_t$, cutting off the link $\mathbf{C}_t \to \mathbf{X}_t$, which gives $\mathbf{X}_t$ a fair opportunity to incorporate each contextual condition $\mathbf{C}$ into spatio-temporal state $\mathbf{H}_t$. Formally, we have:
\begin{equation}\label{equ:front}
    \begin{aligned}
        &P({\mathbf{H}_{t}}|do({\mathbf{X}_{t}})) =\sum\limits_{{\mathbf{S}_{t}},{\mathbf{T}_{t}}}{P({\mathbf{S}_{t}},{\mathbf{T}_{t}}|do({\mathbf{X}_{t}}))}P({\mathbf{H}_{t}}|do({\mathbf{S}_{t}},{\mathbf{T}_{t}})) \\
        & =\sum\limits_{{\mathbf{S}_{t}},{\mathbf{T}_{t}}}{P({\mathbf{S}_{t}},{\mathbf{T}_{t}}|{\mathbf{X}_{t}})}\sum\limits_{{{{\mathbf{X}'}}_{t}}}{P({\mathbf{H}_{t}}|{\mathbf{S}_{t}},{\mathbf{T}_{t}},{{{\mathbf{X}'}}_{t}})}P({{{\mathbf{X}'}}_{t}})\text{ } \\
    \end{aligned}
\end{equation}
where $P({{{\mathbf{X}'}}_{t}})$ represents the prior distribution of the input data, and we propose an \textbf{Input Gate} to fit this prior distribution.
$P({\mathbf{S}_{t}},{\mathbf{T}_{t}}|{\mathbf{X}_{t}})$ denotes the process of extracting spatio-temporal features from data and noise, and we propose a \textbf{Dynamic Causality Generator} to embed spatio-temporal causality into a dynamic causal graph.
$P({\mathbf{H}_{t}}|{\mathbf{S}_{t}},{\mathbf{T}_{t}},{{{\mathbf{X}'}}_{t}})$ denotes the generation of time-varying spatio-temporal states from features to describe the spatio-temporal patterns inherent in the data, and we propose a \textbf{Spatio-temporal Evolutionary Graph Convolution} to extract spatio-temporal states.

\subsubsection{Input Gate}
The periodic flow data are composed of the $P$ slices of bike flow tensor from the previous week ${\mathbf{X}_{t-P+1:t}^{week}}\in {{\mathbb{R}}^{N\times P\times {{c}_{1}}}}$, the previous day ${\mathbf{X}_{t-P+1:t}^{day}}$, and previous time steps ${\mathbf{X}_{t-P+1:t}^{hour}}$, where ${c}_{1}=2$ is the number of flow features. For example, assuming that the time step is one hour, we use the historical four hours ($P = 4$) to predict the bike flow for the next two hours ($Q = 2$) from 18:00-20:00 on Monday. We take the historical periodic flow data as input, including 16:00-20:00 from the previous week, 16:00-20:00 from the previous day and 14:00-18:00 from the previous four hours. Analogously, we can collect the external conditions at the same time step, denoted as ${\mathbf{C}_{t-P+1:t}^{week}}\in {{\mathbb{R}}^{N\times P\times {{c}_{2}}}}$, ${\mathbf{C}_{t-P+1:t}^{day}}$, ${\mathbf{C}_{t-P+1:t}^{hour}}$ respectively, where ${c}_{2}$ is the number of external conditions.

As shown in Fig.\ref{fig:2}, the input gate gets a concatenation of the periodic flow data and the external conditions within the same time step (e.g. $\mathbf{X}_{t}^{day}||\mathbf{C}_{t}^{day}$, where $||$ denotes the concatenation operation). The fully connected layer is used to process the input features of week, day, and hour separately. Then we concatenate them together to obtain $\mathbf{X}_{t}^{in}$. Finally, the gated linear unit is used to output the context-conditioned features ${{\mathbf{{X}}}'_{t}}$. Given the concatenated input features $\mathbf{X}_{t}^{in}$, the gated linear unit is defined as follows:
\begin{equation}\label{equ:5}
    {{\mathbf{{X}}}'_{t}}=\mathbf{X}_{t}^{in}+\phi (\mathbf{X}_{t}^{in}{{\mathbf{\Theta }}_{1}}+\mathbf{a})\odot  \sigma (\mathbf{X}_{t}^{in}{{\mathbf{\Theta }}_{2}}+\mathbf{b}) \in {{\mathbb{R}}^{N\times d}}
\end{equation}
where ${{\mathbf{\Theta }}_{1}}$, ${{\mathbf{\Theta }}_{2}}$, $\mathbf{a}$, and $\mathbf{b}$ are model parameters, $\odot $ is the element-wise product, $\phi (\cdot )$ is the tanh function, and $\sigma (\cdot )$ is the sigmoid function.

\subsubsection{Dynamic Causality Generator}

The conditions of bike-sharing systems are complex, and inter-regional relationships are affected not only by spatio-temporal causality, but also by diverse external conditions. However, most methods related to dynamic graphs focus on the spatial correlations between nodes while ignoring the influence of time-varying contextual conditions and causality. As shown in Fig.\ref{fig:2}, we propose an Dynamic Causality Generator with tightly coupled spatio-temporal states and context-conditioned features.

At each time step, the output of input gate ${{\mathbf{{X}}}'_{t}}$ and spatio-temporal state of the previous time step ${{\mathbf{H}}_{t-1}}$ are concatenated as the input of dynamic causality generator:
\begin{equation}\label{equ:6}
    {\mathbf{I}_{t}}=({{\mathbf{{X}}}'_{t}}||{{\mathbf{H}}_{t-1}}){\mathbf{\Theta }_{dyn}}+{\mathbf{b}_{dyn}}
\end{equation}
where ${\mathbf{I}_{t}}\in {{\mathbb{R}}^{N\times d}}$, ${\mathbf{\Theta }_{dyn}}$ and $\mathbf{b}_{dyn}$ are model parameters, $d$ is the number of feature channels.
We apply the squeeze-excitation method in the node dimension of ${\mathbf{I}_{t}}$ to learn the importance-aware vectorized representation of each node. According to the importance of different nodes, it can promote useful features and suppress features that have little effect on the current task, so that each node can be differentially expressed, which is helpful for subsequent similarity calculation. First, the global channel information is squeezed into the node descriptor using the global average pool:
\begin{equation}\label{equ:7}
    {{\mathbf{z}}_{s}}={{\operatorname{F}}_{sq}}({{\mathbf{I}}_{t}})=\frac{1}{d}\sum\limits_{c=1}^{d}{{{\mathbf{I}}_{t}}[:,c]}\in {{\mathbb{R}}^{N}}
\end{equation}
Second, the specificity of the node is excited by a self-gating mechanism based on node dependence:
\begin{equation}\label{equ:8}
    {{\mathbf{z}}_{e}}={\operatorname{F}_{ex}}({{\mathbf{z}}_{s}})=\sigma ({{\mathbf{\Theta }}_{ex2}}\operatorname{ReLU}({{\mathbf{\Theta }}_{ex1}}{{\mathbf{z}}_{s}}))\in {{\mathbb{R}}^{N}}
\end{equation}
where ${{\mathbf{z}}_{e}}\in {{\mathbb{R}}^{N}}$, ${{\mathbf{\Theta }}_{ex1}}\in {{\mathbb{R}}^{\frac{N}{16}\times N}}$ and ${{\mathbf{\Theta }}_{ex2}}\in {{\mathbb{R}}^{N\times \frac{N}{16}}}$ are model parameters, $\sigma (\cdot )$ is the sigmoid function, and $\operatorname{ReLU}(\cdot )$ is the ReLU function. Then, ${\mathbf{I}_{t}}$ is weighted to generate the dynamic latent representation of each node.
\begin{equation}\label{equ:9}
    \mathbf{DX}_{t}={{\operatorname{F}}_{scale}}({{\mathbf{I}}_{t}},{{\mathbf{z}}_{e}})={{\mathbf{I}}_{t}}\odot {{\mathbf{z}}_{e}}\in {{\mathbb{R}}^{N\times d}}
\end{equation}
Finally, we follow the idea of self-attention to calculate the inter-node similarity and embed the dynamic causality into the causal graph:
\begin{equation}\label{equ:10}
    \mathbf{A}_{t}^{dyn}=\operatorname{ReLU}(\phi (\frac{\mathbf{D}{{\mathbf{X}}_{t}}\mathbf{D}{{\mathbf{X}}_{t}}^{T}}{\sqrt{d}}))\in {{\mathbb{R}}^{N\times N}}
\end{equation}
where $\phi (\cdot )$ is the tanh function.

\subsubsection{Spatio-temporal Evolutionary Graph Convolution}

The static topological graphs and the causal graphs based on bike-sharing systems reflect the inter-node relationship from diverse perspectives. Therefore, we propose a Spatio-temporal Evolutionary Graph Convolution Network (STEGCN) to specify Eq.\ref{equ:front}, as shown in Fig.\ref{fig:2}.

For static topological graphs, we establish the geographical distance graph ${{\mathbf{A}}^{geo}}$ and the transition probability graph ${{\mathbf{A}}^{trans}}$ based on the historical trip records of the regions. Their respective weighted adjacency matrices are as follow:

\begin{small}
\begin{equation}\label{equ:11}
    \begin{matrix}
       \mathbf{A}_{kl}^{geo}=\left\{
       \begin{matrix}
        \exp (-\frac{dis_{kl}^{2}}{{{\sigma }^{2}}}) & di{{s}_{kl}}>{{\varepsilon }^{geo}}  \\
        0 & otherwise  \\
        \end{matrix},\right. & \mathbf{A}_{kl}^{trans}=\frac{{{T}_{kl}}}{\sum\nolimits_{c=1}^{N}{{{T}_{kc}}}}  \\
    \end{matrix}
\end{equation}
\end{small}
where $di{{s}_{kl}}, 1\le k,l\le N$ is distance between region ${r}_{k}$ and region ${r}_{l}$ calculated by the latitude and longitude of the regional center, ${\varepsilon }^{geo}$ denotes distance threshold and is set as 2 kilometers according to the actual situation, $\sigma$ is the variance of the distance matrix, which is used to control the distribution and sparsity of matrix, and ${{T}_{kl}}$ is transition flow.

The STEGCN can be defined as follows:
\begin{equation}\label{equ:12}
    \begin{aligned}
         \mathbf{X}_{t}^{(0)}=&({{\mathbf{X}}'_{t}}||{{\mathbf{H}}_{t-1}})\in {{\mathbb{R}}^{N\times 2d}} \\
         \mathbf{X}_{t}^{(n)}=&{{\alpha }_{0}}\mathbf{X}_{t}^{(n-1)}+{{\alpha }_{1}}{{{\mathbf{\tilde{A}}}}^{geo}}\mathbf{X}_{t}^{(n-1)}+ \\
                              &{{\alpha }_{2}}{{{\mathbf{\tilde{A}}}}^{trans}}\mathbf{X}_{t}^{(n-1)}+{{\alpha }_{3}}\mathbf{\tilde{A}}_{t}^{dyn}\mathbf{X}_{t}^{(n-1)}\in {{\mathbb{R}}^{N\times 2d}} \\
         \mathbf{\tilde{X}}_{t}^{out}=&\operatorname{ReLU}(\sum\limits_{k=0}^{n}{\mathbf{X}_{t}^{(k)}}{{\mathbf{W}}^{(k)}}+{{\mathbf{b}}^{(k)}})\in {{\mathbb{R}}^{N\times d}}
    \end{aligned}
\end{equation}
where ${{\mathbf{\tilde{A}}}^{geo}}$, ${{\mathbf{\tilde{A}}}^{trans}}$ and $\mathbf{\tilde{A}}_{t}^{dyn}$ denotes the normalized adjacency matrices, defined by $\mathbf{\tilde{A}}={{\mathbf{D}}^{-1}}\mathbf{A},{{\mathbf{D}}_{ii}}=\sum\nolimits_{j}{{{\mathbf{A}}_{ij}}}$, $\mathbf{W}^{(k)}$ and ${{\mathbf{b}}^{(k)}}$ are model parameters, ${\alpha }_{0}$, ${\alpha }_{1}$, ${\alpha }_{2}$, ${\alpha }_{3}$ are contribution coefficient that can be learned, and $n$ is the depth of propagation. The contributions of different graphs for bike flow prediction can be learned by training ${\alpha }$. Eq.\ref{equ:12} is the specific implementation of Eq.\ref{equ:front}. The 2nd sum in Eq.\ref{equ:front} corresponds to the 2nd formula in Eq.\ref{equ:12}, indicating that spatio-temporal features are extracted from different aspects (geographic, transition, causality) by means of graph convolution, and they are weighted to sum (i.e. ${P({\mathbf{S}_{t}},{\mathbf{T}_{t}}|{\mathbf{X}_{t}})}\sum{P({\mathbf{H}_{t}}|{\mathbf{S}_{t}},{\mathbf{T}_{t}},{{{\mathbf{X}'}}_{t}})}P({{{\mathbf{X}'}}_{t}})$). The 1st sum in Eq.\ref{equ:front} corresponds to the last formula of Eq.\ref{equ:12}, indicating that all the above information is combined (i.e. $\sum{P({\mathbf{S}_{t}},{\mathbf{T}_{t}}|{\mathbf{X}_{t}})}\sum{P({\mathbf{H}_{t}}|{\mathbf{S}_{t}},{\mathbf{T}_{t}},{{{\mathbf{X}'}}_{t}})}P({{{\mathbf{X}'}}_{t}})$).

We simplify Eq.\ref{equ:12} to $\mathbf{\Theta }{{\star }_{G}}({{\mathbf{{X}}}'_{t}}||{{\mathbf{H}}_{t-1}})$. Specifically, we adopt diffusion convolution\cite{li2018dcrnn} to separately propagate the inflow and outflow information of each node in the directed graph, represented as follows:
\begin{equation}\label{equ:13}
    \begin{aligned}
        \mathbf{X}_{t}^{out}=&\mathbf{\Theta }{{\star }_{G}}({{\mathbf{{X}}}'_{t}}||{{\mathbf{H}}_{t-1}}) \\
                            =&{{\mathbf{\Theta }}_{1}}{{\star }_{G}}(\mathbf{X}_{t}^{(0)},\mathbf{A })+{{\mathbf{\Theta }}_{2}}{{\star }_{G}}(\mathbf{X}_{t}^{(0)},{{\mathbf{A}}^{T}})
    \end{aligned}
\end{equation}

\subsubsection{Spatio-temporal Neural Structural Causal Unit}

We integrate the input gate, dynamic causality generator and spatio-temporal evolutionary graph convolution derived from the frontdoor criterion into the Spatio-temporal Neural Structural Causal Unit (STNSCU) to represent the complete causal intervention process $P({\mathbf{H}_{t}}|do({\mathbf{X}_{t}}))$. As shown in Fig.\ref{fig:2}, STNSCU can effectively model complex nonlinear spatio-temporal causality, denoted as follows:
\begin{equation}\label{equ:14}
    \begin{aligned}
       {{\mathbf{r}}_{t}}&=\sigma ({{\mathbf{\Theta }}_{r}}{{\star }_{G}}({{{\mathbf{{X}}}}'_{t}}||{{\mathbf{H}}_{t-1}})+{{\mathbf{b}}_{r}}) \\
      {{\mathbf{z}}_{t}}&=\sigma ({{\mathbf{\Theta }}_{z}}{{\star }_{G}}({{{\mathbf{{X}}}}'_{t}}||{{\mathbf{H}}_{t-1}})+{{\mathbf{b}}_{z}}) \\
      {{{\mathbf{\tilde{h}}}}_{t}}&=\phi ({{\mathbf{\Theta }}_{h}}{{\star }_{G}}({{{\mathbf{{X}}}}'_{t}}||({{\mathbf{r}}_{t}}\odot {{\mathbf{H}}_{t-1}}))+{{\mathbf{b}}_{h}}) \\
      {{\mathbf{H}}_{t}}&={{\mathbf{z}}_{t}}\odot {{\mathbf{H}}_{t-1}}+(1-{{\mathbf{z}}_{t}})\odot {{{\mathbf{\tilde{h}}}}_{t}}
    \end{aligned}
\end{equation}
where ${\mathbf{\Theta }}_{r}$, ${\mathbf{\Theta }}_{z}$, ${\mathbf{\Theta }}_{h}$, ${\mathbf{b}}_{r}$, ${\mathbf{b}}_{z}$, and ${\mathbf{b}}_{h}$ are the parameters of graph convolution, ${\star }_{G}$ is graph convolution defined by Eq.\ref{equ:13}, and ${{\mathbf{H}}_{t}}$ is the spatio-temporal state of the STNSCU at time step $t$.

In the multi-step forecasting model, STNSCM is an encoder-decoder structure composed of STNSCUs. Historical periodic flow data and external conditions are fed into the encoder and predictions ${{\mathbf{\hat{X}}}^{pred}}={{\mathbf{\hat{X}}}_{t+1:t+Q}}\in {{\mathbb{R}}^{N\times Q\times {{c}_{1}}}}$ are output by the decoder. The spatio-temporal states of the encoder are transformed to initialize the decoder with a counterfactual representation reasoning module. Since the future external conditions ${{\mathbf{C}}_{t+1:t+Q}}\in {{\mathbb{R}}^{N\times Q\times {{c}_{2}}}}$ are accessible, it is concatenated with the previous prediction as part of the decoder input.

\subsection{Counterfactual Representation Reasoning}

Under the condition of $\mathbf{X}_t$ in the factual scenario, $\mathbf{C}$ is set to $\mathbf{C}_{t+1}$ to infer the spatio-temporal state $\mathbf{H}'_t$ in the counterfactual scenario, and then predict the future bike flow $\mathbf{Y}_{t+1}$ via $\mathbf{H}'_t$. Formally, we have:
{\scriptsize
\begin{equation}\label{equ:counterfactual}
    \begin{aligned}
        P \! \left(\mathbf{Y}_{C_{t+1}} \! \mid \! \mathbf{X}_{t}\right) \! &= \! \sum_{\mathbf{H}_{t}^{\prime}} \!P \!\left(\mathbf{Y}_{t+1} \! \mid \! d o \! \left(\mathbf{H}_{t}^{\prime}\right) \! , \! do\left(\mathbf{C}_{t+1}\right) \! , \! \mathbf{X}_{t}\right) \! P \! \left(\mathbf{H}_{t}^{\prime} \! \mid \! \mathbf{X}_{t} \! , d o \! \left(\mathbf{C}_{t+1}\right)\right) \\
        &=\sum_{\mathbf{H}_{t}^{\prime}} P\left(\mathbf{H}_{t}^{\prime} \mid \mathbf{X}_{t}, C_{t+1}\right) P\left(\mathbf{Y}_{t+1} \mid \mathbf{H}_{t}^{\prime}, \mathbf{C}_{t+1}\right)
    \end{aligned}
\end{equation}}
where $P\left(\mathbf{H}_{t}^{\prime} \mid \mathbf{X}_{t}, \mathbf{C}_{t+1}\right)$ represents the counterfactual representation reasoning process, and $P\left(\mathbf{Y}_{t+1} \mid \mathbf{H}_{t}^{\prime}, \mathbf{C}_{t+1}\right)$ represents the prediction process based on counterfactual representation.

Our purpose is to infer the spatio-temporal state in the case of $\mathbf{C}_{t+1}$ by focusing on the similar part between the external conditions of the future and history. Counterfactual representation reasoning module employs scaled dot-product attention to dynamically calculate relationships between each future and historical time step, and converts the encoded historical spatio-temporal states to future representations. Finally, future representations are used to initialize the decoder. As shown in Fig.\ref{fig:2}, a fully connected layer is used to generate future external features ${{\mathbf{F}}^{pred}}=\operatorname{FC}\left( {{\mathbf{C}}_{t+1:t+Q}} \right)\in {{\mathbb{R}}^{N\times Q\times d}}$ and historical external features ${{\mathbf{C}}^{his}}=\operatorname{FC}\left( {{\mathbf{C}}^{week}}||{{\mathbf{C}}^{day}}||{{\mathbf{C}}^{hour}} \right)\in {{\mathbb{R}}^{N\times P\times d}}$, which are taken as the query and key of the attention mechanism. The historical spatio-temporal states ${{\mathbf{H}}^{his}}={{\mathbf{H}}_{t-P+1:t}}\in {{\mathbb{R}}^{N\times P\times d}}$ are taken as the value. The counterfactual representation reasoning module is formulated as:
\begin{equation}\label{equ:15}
    \begin{aligned}
       \mathbf{Q}&={{\mathbf{C}}^{pred}}{{\mathbf{W}}^{Q}}\in {{\mathbb{R}}^{N\times Q\times d}} \\
       \mathbf{K}&={{\mathbf{C}}^{his}}{{\mathbf{W}}^{K}}\in {{\mathbb{R}}^{N\times P\times d}} \\
       \mathbf{V}&={{\mathbf{H}}^{his}}{{\mathbf{W}}^{V}}\in {{\mathbb{R}}^{N\times P\times d}} \\
       {{\mathbf{H}}^{pred}}&=\operatorname{softmax}(\frac{\mathbf{Q}{{\mathbf{K}}^{T}}}{\sqrt{d}})\mathbf{V}\in {{\mathbb{R}}^{N\times Q\times d}} \\
    \end{aligned}
\end{equation}
where $d$ is the number of feature channels. ${\mathbf{W}}^{Q}$, ${\mathbf{W}}^{K}$, and ${\mathbf{W}}^{V}$ are learnable parameters.
Intuitively, for historical spatio-temporal states, the counterfactual representation reasoning module indicates to pay more attention to the parts whose external conditions are similar to the future. The future representations ${{\mathbf{H}}^{pred}}$ are input into a fully connected layer and then used to initialize the decoder.

\section{Experiments}
In this section, we evaluate the effectiveness of STNSCM by experiments conducted on real-world datasets\footnote{\url{https://github.com/EternityZY/STNSCM}}.

\subsection{Experimental Settings}

\textbf{Datasets}: We collect two real-world datasets, NYC-Bike and BJ-Bike, each dataset contains the corresponding weather and time information. We split the dataset with a 30-minute interval. We select the first 60\% of data as the training set, 20\% as the validation set, and 20\% as the test set.


%

\textbf{Baselines}: We compare STNSCM with recent state-of-the-art baselines. We summarize the models into four categories, including deep learning methods, predefined graph methods, adaptive graph methods, attention methods and dynamic graph methods. The main difference between adaptive graph and dynamic graph is whether the change of graph structure depends on input features.

\textbf{Evaluation Metrics}: We evaluate the performance of methods with Root Mean Square Error (RMSE), Mean Absolute Error (MAE) and Mean Absolute Percentage Error (MAPE).


\subsection{Comparison with Baselines}

\begin{table*}[]
\centering
\caption{Performance comparison with other models.}
\label{tab:2}
\resizebox{\linewidth}{!}{
\begin{tabular}{ccc|ccc|ccc|ccc}
\hline
\multirow{2}{*}{dataset}   & \multirow{2}{*}{Category}                                                   & \multirow{2}{*}{Models} & \multicolumn{3}{c|}{30min}                               & \multicolumn{3}{c|}{60min}                               & \multicolumn{3}{c}{Avg}                                  \\ \cline{4-12}
                           &                                                                             &                         & MAE              & RMSE             & MAPE               & MAE              & RMSE             & MAPE               & MAE              & RMSE             & MAPE               \\ \hline
\multirow{11}{*}{BJ-Bike}  & \multirow{2}{*}{\begin{tabular}[c]{@{}c@{}}Deep\\ Learning\end{tabular}}    & LSTM                    & 14.2031          & 28.2169          & 19.5308\%          & 19.8906          & 41.4004          & 26.1706\%          & 17.378           & 35.3137          & 22.9824\%          \\
                           &                                                                             & GRU                     & 14.5063          & 30.5848          & 20.1777\%          & 20.2897          & 42.8662          & 27.1207\%          & 17.7293          & 37.0763          & 23.7914\%          \\ \cline{2-12}
                           & \multirow{2}{*}{\begin{tabular}[c]{@{}c@{}}Predefined\\ Graph\end{tabular}} & STGCN                   & 11.5225          & 32.3062          & 15.4271\%          & 14.1583          & 36.1548          & 18.3616\%          & 13.2019          & 33.4340          & 16.9863\%          \\
                           &                                                                             & STGODE                  & 13.0722          & 25.5082          & 17.6931\%          & 18.2863          & 38.2222          & 23.7129\%          & 15.9745          & 32.0672          & 20.8288\%          \\ \cline{2-12}
                           & \multirow{4}{*}{\begin{tabular}[c]{@{}c@{}}Adaptive\\ Graph\end{tabular}}   & GWNet                   & 11.3790          & 23.4735          & 15.7683\%          & 14.4496          & 30.3216          & 20.0026\%          & 13.0956          & 25.7935          & 17.9991\%          \\
                           &                                                                             & HGCN                    & 12.3808          & 23.7278          & 16.5783\%          & 14.4550          & 29.5130          & 19.1782\%          & 13.6700          & 25.9349          & 18.0185\%          \\
                           &                                                                             & CCRNN                   & 13.0028          & 40.7625          & 16.2663\%          & 15.1600          & 43.8774          & 19.0467\%          & 14.4291          & 40.7677          & 17.7535\%          \\
                           &                                                                             & DMSTGCN                 & 11.3967          & 23.1091 & 15.6620\%          & 14.1286          & 29.5651          & 18.9005\%          & 12.9
                           921          & 25.6640          & 17.3526\%          \\ \cline{2-12}
                           & \multirow{2}{*}{\begin{tabular}[c]{@{}c@{}}Attention\\ Graph\end{tabular}}  & GMAN                    & 14.2979          & 32.5408          & 19.4858\%          & 19.5415          & 43.8546          & 25.7455\%          & 17.3069          & 38.4888          & 22.7454\%          \\
                           &                                                                             & ASTGNN                  & 13.0494          & 26.2419          & 17.4269\%          & 17.8318          & 40.4308          & 23.1511\%          & 15.8104          & 33.4645          & 20.4270\%          \\ \cline{2-12}
                           & \multirow{2}{*}{\begin{tabular}[c]{@{}c@{}}Dynamic\\ Graph\end{tabular}}    & DGCRN                   & 11.4163          & 27.7059          & 16.0872\%          & 14.0468          & 33.2321          & 19.2002\%          & 12.9966          & 29.8171          & 17.7582\%          \\
                           &                                                                             & \textbf{STNSCM}         & \textbf{11.2833}         & \textbf{23.2897}          & \textbf{15.2978\%} & \textbf{13.3415} & \textbf{28.1254} & \textbf{17.4721\%} & \textbf{12.5180} & \textbf{24.4383} & \textbf{16.4879\%} \\ \hline
\multirow{11}{*}{NYC-Bike} & \multirow{2}{*}{\begin{tabular}[c]{@{}c@{}}Deep\\ Learning\end{tabular}}    & LSTM                    & 3.1259           & 6.0591           & 24.8426\%          & 3.8342           & 7.9119           & 30.9824\%          & 3.4809           & 6.9045           & 28.5997\%          \\
                           &                                                                             & GRU                     & 3.1359           & 6.0629           & 24.8762\%          & 3.8442           & 7.8694           & 30.9406\%          & 3.4908           & 6.8827           & 28.6149\%          \\ \cline{2-12}
                           & \multirow{2}{*}{\begin{tabular}[c]{@{}c@{}}Predefined\\ Graph\end{tabular}} & STGCN                   & 2.6015           & 4.9071           & 20.5065\%          & 2.9732           & 6.1072           & 23.3678\%          & 2.7879           & 5.4049           & 22.4650\%          \\
                           &                                                                             & STGODE                  & 2.7222           & 5.2398           & 21.4485\%          & 3.2079           & 6.7118           & 25.4689\%          & 2.9649           & 5.8408           & 23.8943\%          \\ \cline{2-12}
                           & \multirow{4}{*}{\begin{tabular}[c]{@{}c@{}}Adaptive\\ Graph\end{tabular}}   & GWNet                   & 2.5686           & 4.8377           & 20.5743\%          & 2.9589           & 6.1023           & 23.7937\%          & 2.7644           & 5.3748           & 22.6851\%          \\
                           &                                                                             & HGCN                    & 2.5865           & 4.9755           & 20.2884\%          & 2.9728           & 6.4410           & 23.4383\%          & 2.7803           & 5.5742           & 22.3060\%          \\
                           &                                                                             & CCRNN                   & 2.5945           & 4.8986           & 20.4141\%          & 2.9670           & 6.2435           & 23.4624\%          & 2.7814           & 5.4886           & 22.4958\%          \\
                           &                                                                             & DMSTGCN                 & 2.5539           & 4.7400           & 20.1860\%          & 2.9159           & 6.0370           & 23.2154\%          & 2.7395           & 5.3201           & 22.1471\%          \\ \cline{2-12}
                           & \multirow{2}{*}{\begin{tabular}[c]{@{}c@{}}Attention\\ Graph\end{tabular}}  & GMAN                    & 3.1153           & 6.2649           & 23.6753\%          & 3.1816           & 6.4593           & 24.7447\%          & 3.1469           & 6.1648           & 24.7131\%          \\
                           &                                                                             & ASTGNN                  & 2.9774           & 5.6568           & 23.3848\%          & 3.3349           & 6.8282           & 26.2812\%          & 3.1576           & 6.0232           & 25.3607\%          \\ \cline{2-12}
                           & \multirow{2}{*}{\begin{tabular}[c]{@{}c@{}}Dynamic\\ Graph\end{tabular}}    & DGCRN                   & 2.6175           & 4.9426           & 20.5837\%          & 2.9653           & 6.1419           & 23.3420\%          & 2.7918           & 5.4210           & 22.4856\%          \\
                           &                                                                             & \textbf{STNSCM}         & \textbf{2.5289}  & \textbf{4.6835}  & \textbf{19.8475\%} & \textbf{2.8099}  & \textbf{5.6129}  & \textbf{22.4450\%} & \textbf{2.6701}  & \textbf{5.0397}  & \textbf{21.5300\%} \\ \hline
\end{tabular}}
\end{table*}


For fairness, we deploy the same environment, loss function, periodic flow data, and external conditions for all models. We compare STNSCM with baselines for traffic prediction and the final average results are shown in Table \ref{tab:2}.

We classify these methods according to whether the graph structure varies depending on input features. Results show our STNSCM outperforms baseline models consistently and overwhelmingly. In deep learning methods, Poor performances of indicate the limitation of failing to consider spatial correlation. STGODE\cite{KDD2021ODEGCN} deepens networks to extract higher-order features through a tensor-based ordinary differential equation, but hampered by the amount of data, STGODE shows a worse performance. Besides, these methods only rely on the fixed graph structure while ignoring the dynamic characteristics.

The adaptive graph\cite{graphwavenet2019, AAAI2021HGCN, ye2021CGCdemand, KDD2021DMSTGCN} is helpful in the short-term prediction (30min), but it is still static over time and fails to capture time-varying spatio-temporal dependencies, and the effect of long-term prediction (60min) is significantly reduced.
The attention graph\cite{AAAI2020GMAN, TKDE2021ASTGNN} can only change the weight of predefined graphs rather than the structure.
DGCRN\cite{li2021DGCRN} employs the pre-defined adjacency matrix to conduct the message-passing process for dynamic node status. Due to the incomplete connections in the data, the predefined graph itself may contain noise, hindering the generation of dynamic graphs.

Besides, these models do not have exclusive modules to handle contextual conditions, and we merely concatenate them with periodic flow data, so external information is not fully exploited. The qualitative prediction results are shown in Fig. \ref{fig:1}(e). The main contribution of our model is stability and resistance to random fluctuations, which are also the most important capabilities of bike flow forecasting. Therefore, in the NYC-Bike test set, the bike flow distribution fluctuates greatly due to the external conditions, so our method outperforms all methods.


\subsection{Component Analysis}

\begin{table}[]
\centering
\caption{Comparison with variants of STNSCM on NYC-Bike dataset.}
\label{tab:3}
\resizebox{\linewidth}{!}{
\begin{tabular}{ccccc}
\hline
\multirow{2}{*}{Category}                                                                 & \multirow{2}{*}{Models} & \multicolumn{3}{c}{Average}                            \\ \cline{3-5}
                                                                                          &                         & MAE             & RMSE            & MAPE               \\ \hline
\multirow{3}{*}{Graph}                                                                    & w/o ${{\mathbf{A}}^{geo}}$                & 2.7101          & 5.2185          & 21.8458\%          \\
                                                                                          & w/o ${{\mathbf{A}}^{trans}}$              & 2.7124          & 5.2515          & 22.0781\%          \\
                                                                                          & w/o ${{\mathbf{A}}^{dyn}}$                & 2.7411          & 5.3144          & 22.3316\%          \\ \hline
\multirow{3}{*}{\begin{tabular}[c]{@{}c@{}}Dynamic\\ Causality\\ Generator\end{tabular}} & EGG w/o SE              & 2.7074          & 5.1928          & 21.8821\%          \\
                                                                                          & EGG w/o $H$             & 2.7300          & 5.3132          & 22.2931\%          \\
                                                                                          & EGG w/o $X$             & 2.7105          & 5.2612          & 22.0250\%          \\ \hline
\multirow{2}{*}{\begin{tabular}[c]{@{}c@{}}Input\\ Gate\end{tabular}}
& IG w/ FC                 & 2.7266          & 5.3050          & 22.2241\%          \\
& w/o IG                  & 2.7239          & 5.3355          & 22.3176\%          \\ \hline
\begin{tabular}[c]{@{}c@{}}Counterfactual\\
Representation\end{tabular}                        & w/o CR                  & 2.7041          & 5.2094          & 21.8430\%          \\ \hline
                                                                                          & STNSCM                  & \textbf{2.6701} & \textbf{5.0397} & \textbf{21.5300\%} \\ \hline
\end{tabular}}
\end{table}

\begin{figure}
  \centering
  \includegraphics[width=0.4\textwidth]{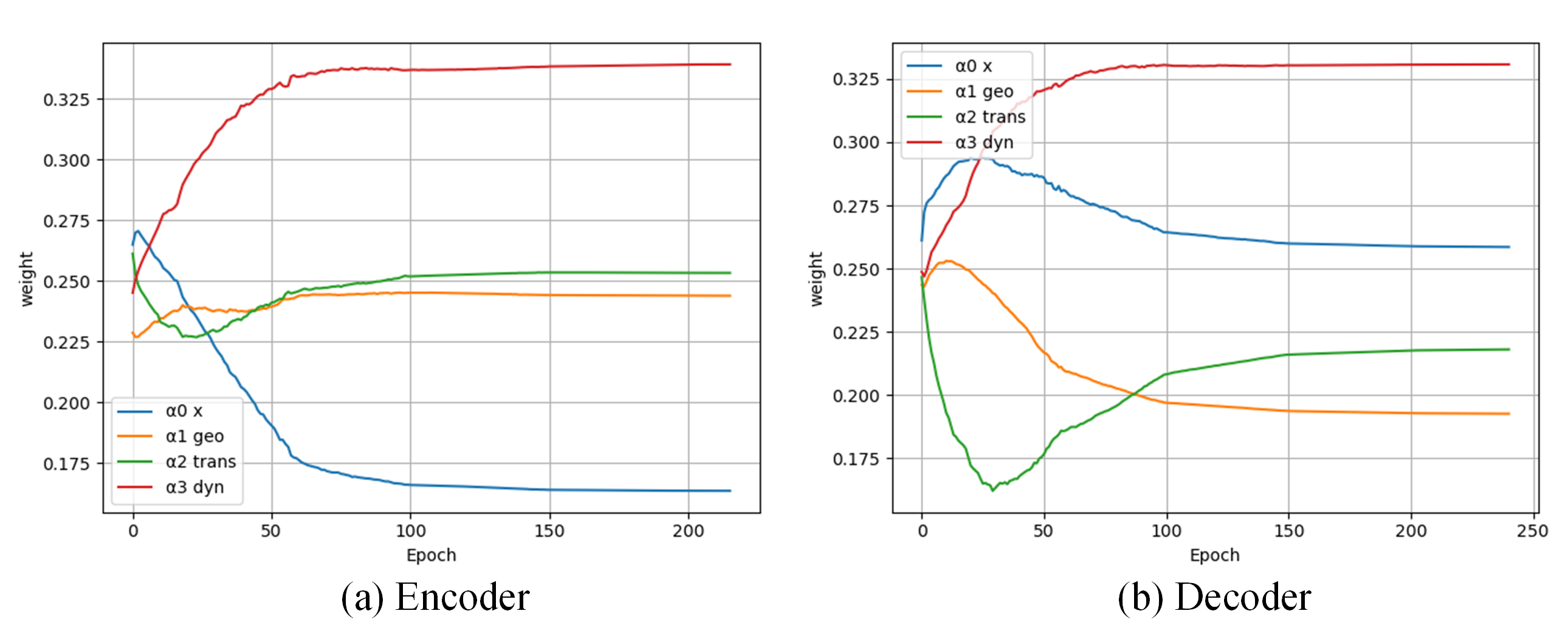}
  \caption{The change process of the contribution coefficient $\alpha$ in the spatio-temporal evolutionary graph convolution during the training period.}
  \label{fig:7}
\end{figure}

\begin{figure}
  \centering
  \includegraphics[width=0.45\textwidth]{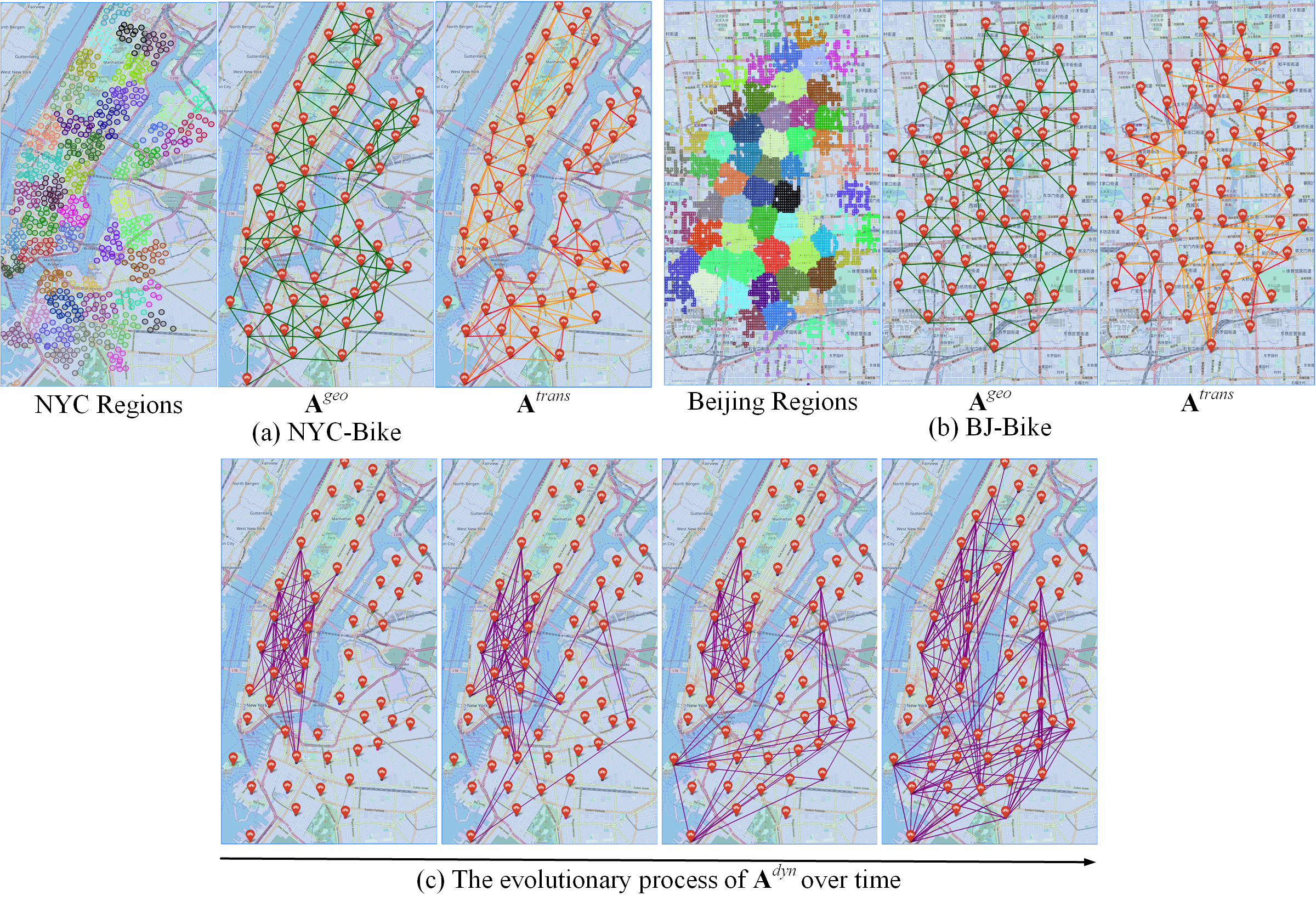}
  \caption{Region-based static topologies and dynamic causal graphs.}
  \label{fig:5}
\end{figure}

To verify effectiveness of key components in STNSCM, we conduct ablation experiments on NYC-Bike dataset, which are described as follows:
\begin{itemize}
\item w/o ${{\mathbf{A}}^{geo}}$: It removes the geographical distance graph.
\item w/o ${{\mathbf{A}}^{trans}}$: It removes the transition probability graph.
\item w/o ${{\mathbf{A}}^{dyn}}$: It removes the dynamic causal graph.
\item EGG w/o SE: It removes the squeeze-excitation method from dynamic causality generator.
\item EGG w/o $\mathbf{H}$: It removes the spital-temporal state of the previous time step from dynamic causality generator.
\item EGG w/o $\mathbf{X}$: It removes the features output by input gate from dynamic causality generator.
\item w/o CR: It removes the counterfactual representation reasoning module from STNSCM. The last spital-temporal state of the encoder are copied to initialize the decoder.
\item IG w/ FC: It concatenates periodic flow data and external conditions to extract context-conditioned features only through fully connected layers.
\item w/o IG: It removes the input gate. The model merely input the bike flow tensor of the previous $P$ time steps.
\end{itemize}

The performance of all variants is summarized in Table \ref{tab:3}. For the contribution of different graphs, the dynamic causal graph obviously plays a more prominent role. The introduction of the causal graph can significantly improve performance, as it provides implicit causality that cannot be extracted from static topological graphs. Meanwhile, the geographical distance graph and the transition probability graph are also necessary. The dynamic causal graph can collaborate with static topological graphs to better model complex transportation systems. We further visualized the change process of the contribution coefficient $\alpha$ in the spatio-temporal evolutionary graph convolution during the training period, as shown in Fig.\ref{fig:7}. In the encoder, the geographic distance graph and the transition probability graph have the same importance, while the input feature $\mathbf{X}_{t}^{(n-1)}$ of the graph convolution has a smaller coefficient. This indicates that the encoder needs to extract potential spatio-temporal dependencies in the transportation system by propagating and aggregating the node features from multi-view graphs. On the contrary, in the decoder, the input features $\mathbf{X}_{t}^{(n-1)}$ of the graph convolution account for a large proportion, which shows that the decoder needs to restore high-level spatio-temporal features to predict the future flow.

For the dynamic causality generator, the spatio-temporal state of the previous time step is crucial for modeling the dynamic spatio-temporal causality into the causal graph. We visualized the dynamic causal graph, as shown in Fig.\ref{fig:5}(c), the dynamic causal graph integrated with periodic features and external conditions directly model the interactive evolutionary process across regions.

In addition, external conditions dominate the model performance, otherwise, the model cannot govern the fluctuations caused by this factor. On the one hand, the performance of IG w/ FC and w/o IG is almost the same. This indicates that features extracted by our input gate are more effective. On the other hand, the model's resistance to random fluctuations can be reflected in MAPE. Removing the input gate will greatly increase MAPE. The input gate can effectively fuse periodic flow data and external conditions to extract the context-conditioned features, which is also the basis for STNSCM to resist fluctuations caused by external environment.

The counterfactual representation reasoning module utilizes the attention mechanism to convert the encoded historical spatio-temporal states to future representations, which reduces information loss and improves the overall performance of the model.

\section{Conclusion}

In this work, we build a causal graph to describe the traffic prediction problem from a perspective of causality.
It shows that due to the disturbance of incomplete observation, there are spurious correlations in the feature extraction process, resulting in the model can only perform general scenarios, but failing in special scenarios.
We propose a novel spatio-temporal neural structural causal model that decomposes the frontdoor criterion into multiple sub-terms and proposes well-designed modules to model these sub-terms. Among them, the dynamic causality generator is the most important. It embeds the inter-regional time-varying causal relationship into the dynamic causal graph, which enables the model to capture the dynamic rules. Second, we propose a counterfactual representation reasoning module, which makes spatio-temporal states in the current factual scenarios have the ability to represent counterfactuals. Detailed experiments and analyses demonstrated the superiority of STNSCM over both classical and state-of-the-art prediction methods.

%

\bibliography{reference}

\end{document}